\documentclass[11pt,a4paper]{article}
\usepackage[hyperref]{naaclhlt2018}
\usepackage{times}
\usepackage{latexsym}
\usepackage{helvet}
\usepackage{courier}
\usepackage{footnote}
\usepackage{graphicx} 
\usepackage{balance}  
\usepackage{booktabs} 
\usepackage{ccicons}  
\usepackage{ragged2e} 
\usepackage{multirow}
\usepackage{multirow}
\usepackage{enumerate}
\usepackage{mathpazo}
\usepackage{enumitem,kantlipsum}
\usepackage{tablefootnote}
\usepackage{url}
\aclfinalcopy 

\title{Detecting Egregious Conversations between \\Customers and Virtual Agents}

\author{Tommy Sandbank, \\\textbf{Michal Shmueli-Scheuer,}\\\textbf{Jonathan Herzig, David Konopnicki}\\
  IBM Research\\
  Haifa 31905, Israel \\
  {\{tommy,shmueli,hjon,davidko\}@il.ibm.com} \\\And
John Richards, \\\textbf{David Piorkowski} \\
  IBM Research \\
  Yorktown Heights, NY USA \\
  {ajtr@us.ibm.com,} \\
	{david.piorkowski@ibm.com}
}
\date{}

\begin{document}
\maketitle
\begin{abstract}
Virtual agents are becoming a prominent channel of interaction in customer service. Not all customer interactions are smooth, however, and some can become almost comically bad. In such instances, a human agent might need to step in and salvage the conversation. Detecting bad conversations is important since disappointing customer service may threaten customer loyalty and impact revenue. In this paper, we outline an approach to detecting such \textit{egregious} conversations, using behavioral cues from the user, patterns in agent responses, and user-agent interaction. Using logs of two commercial systems, we show that using these features improves the detection F1-score by around $20\%$ over using textual features alone. In addition, we show that those features are common across two quite different domains and, arguably, universal. 
\end{abstract}

\section{Introduction}
\label{introduction}

Automated conversational agents (chatbots) are becoming widely used for various tasks such as personal assistants or as customer service agents. Recent studies project that $80\%$ of businesses plan to use chatbots by 2020\footnote{http://read.bi/2gU0szG}, and that chatbots will power $85\%$ of customer service interactions by the year $2020$\footnote{http://gtnr.it/2z428RS}. This increasing usage is mainly due to advances in artificial intelligence and natural language processing~\cite{Hirschberg261} along with increasingly capable chat development environments, leading to improvements in conversational richness and robustness.

Still, chatbots may behave extremely badly, leading to conversations so off-the-mark that only a human agent could step in and salvage them. Consequences of these failures may include loss of customer goodwill and associated revenue, and even exposure to litigation if the failures can be shown to include fraudulent claims. Due to the increasing prevalence of chatbots, even a small fraction of such egregious\footnote{Defined by the dictionary as outstandingly bad.} conversations could be problematic for the companies deploying chatbots and the providers of chatbot services.

In this paper we study detecting these egregious conversations that can arise in numerous ways. For example, incomplete or internally inconsistent training data can lead to false classification of user intent. Bugs in dialog descriptions can lead to dead ends. Failure to maintain adequate context can cause chatbots to miss anaphoric references. In the extreme case, malicious actors may provide heavily biased (e.g., the Tay chatbot\footnote{http://bit.ly/2fwYaa5}) or even hacked misbehaviors.

In this article, we focus on customer care systems. In such setting, a conversation usually becomes egregious due to a combination of the aforementioned problems. The resulting customer frustration may not surface in easily detectable ways such as the appearance of all caps, shouting to a speech recognizer, or the use of profanity or extreme punctuation. Consequently, the chatbot will continue as if the conversation is proceeding well, usually leading to conversational breakdown.

Consider, for example, the anonymized but representative conversation depicted in Figure~\ref{fig:example}. Here the customer aims to understand the details of a flight ticket. In the first two turns, the chatbot misses the customer's intentions, which leads to the customer asking ``Are you a real person?''. The customer then tries to explain what went wrong, but the chatbot has insufficient exposure to this sort of utterance to provide anything but the default response (``I'm not trained on that''). The response seems to upset the customer and leads to a request for a human agent, which is rejected by the system (``We don't currently have live agents''). Such rejection along with the previous responses could lead to customer frustration~\cite{amsel_1992}.

\begin{figure}[t]
  {\footnotesize
	\setlength{\fboxsep}{3pt}
	\fbox{
		\parbox{\columnwidth}{

    Customer: \textit{I got 2 quotes for the flight ticket, but i'm wondering what the details of each ticket are?}

    Agent: \textit{Please select "Buy" next to the ticket you'd like to purchase.  }

    Customer: \textit{No, I don't want to buy yet till I know the details of the flights.}

    Agent: \textit{If you're in the process of renting a car, please continue with by clicking "Next" }

    Customer: \textit{Are you a real person?}

    Agent: \textit{I am a digital assistant. I've been trained to answer questions about travels. Ask me any questions you have. }

    Customer: \textit{I asked a specific question and you gave me a random answer about car rental which I'm not interested in.}

    Agent: \textit{I'm not trained on that yet, but I'm still learning. You may want to rephrase your question and try again.  }

    Customer: \textit{This service is pointless , can i talk to a real live person?}

    Agent: \textit{We don't currently have live agents to chat with online. }
		}}
  }
\caption{Example of an \textit{egregious} conversation.}

	\label{fig:example}
\end{figure}

Being able to automatically detect such conversations, either in real time or through log analysis, could help to improve chatbot quality. If detected in real time, a human agent can be pulled in to salvage the conversation. As an aid to chatbot improvement, analysis of egregious conversations can often point to problems in training data or system logic that can be repaired. While it is possible to scan system logs by eye, the sheer volume of conversations may overwhelm the analyst or lead to random sampling that misses important failures. If, though, we can automatically detect the worst conversations (in our experience, typically under $10\%$ of the total), the focus can be on fixing the worst problems.

Our goal in this paper is to study conversational features that lead to egregious conversations. Specifically, we consider customer inputs throughout a whole conversation, and detect cues such as rephrasing, the presence of heightened emotions, and queries about whether the chatbot is a human or requests to speak to an actual human. In addition, we analyze the chatbot responses, looking for repetitions (e.g. from loops that might be due to flow problems), and the presence of "not trained" responses. Finally, we analyze the larger conversational context exploring, for example, where the presence of a "not trained" response might be especially problematic (e.g., in the presence of strong customer emotion).

The main contributions of this paper are twofold:
(1) This is the first research focusing on detecting egregious conversations in conversational agent (chatbot) setting and
(2) this is the first research using unique agent, customer, and customer-agent interaction features to detect egregiousness.

The rest of this paper is organized as follows. We review related work, then we formally define the methodology for detecting egregious conversations. We describe our data, experimental setting, and results. We then conclude and suggest future directions.

\section{Related Work}
\label{related}

Detecting egregious conversations is a new task, however, there is related work that aim at measuring the general quality of the interactions in conversational systems. These works studied the complementary problem of detecting and measuring user satisfaction and engagement. Early work by~\cite{walkerParadis97,Walker:2001} discussed a framework that maximizes the user satisfaction by considering measures such as number of inappropriate utterances, recognition rates, number of times user requests repetitions, number of turns per interaction, etc. Shortcomings of this approach are discussed by~\cite{Hajdinjak:2006}. Other works focus on predicting the user engagement in such systems. Examples include~\cite{Kiseleva2016,Kiseleva2016sigir,JiangAJOZKK15}. Specifically, these works evaluated chat functionality by asking users to make conversations with an intelligent agent and measured the user satisfaction along with other features such as the automatic speech recognition (ASR) quality and intent classification quality. In
~\cite{Sandbank2017} the authors presented a conversational system enhanced with emotion analysis, and suggested using emotions as triggers for human escalation. In our work, we likewise use emotion analysis as predictive features for egregious conversation. The works of~\cite{Sarikaya2017,sano-kaji-sassano:2017:SIGDIAL} studied reasons why users reformulated utterances in such systems. Specifically, in~\cite{Sarikaya2017} they reported on how the different reasons affect the users' satisfaction. In~\cite{sano-kaji-sassano:2017:SIGDIAL} they focused on how to automatically predict the reason for user's dissatisfaction using different features. Our work also explores user reformulation (or rephrasing) as one of the features to predict egregious conversations. We build on the previous work by leveraging some of the approaches in our classifier for egregious conversations.
In~\cite{WalkerWL00,HastiePW02} the authors also looked for problems in a specific setting of spoken conversations. The main difference with our work is that we focus on chat logs for domains for which the expected user utterances are a bit more diverse, using interaction features as well as features that are not sensitive to any architectural aspects of the conversational system (e.g., ASR component).
Several other approaches for evaluating chatbot conversations indirectly capture the notion of conversational quality. For example, several prior works borrowed from the field of pragmatics in various metrics around the principles of cooperative conversation~\cite{Chakrabarti2013AFF,Saygin:2002}. In~\cite{Steidl2004} they measured dialogue success at the turn level as a way of predicting the success of a conversation as a whole. \cite{Webb10evaluatinghuman-machine} created a measure of dialogue appropriateness to determine its role in maintaining a conversation. Recently,~\cite{LiuLSNCP16} evaluated a number of popular measures for dialogue response generation systems and highlighted specific weaknesses in the measures. Similarly, in~\cite{Moller2009} they developed a taxonomy of available measures for an end-user's quality of experience for multimodel dialogue systems, some of which touch on conversational quality. All these measures may serve as reasons for a conversation turning egregious, but none try to capture or predict it directly.

In the domain of customer service, researchers mainly studied reasons for failure of such systems along with suggestions for improved design~\cite{BENMIMOUN2012605,Gnewuch2017}. In~\cite{BENMIMOUN2012605} the authors analyzed reasons sales chatbots fail by interviewing chatbots experts. They found that a combination of exaggerated customer expectations along with a reduction in agent performance (e.g., failure to listen to the consumer, being too intrusive) caused customers to stop using such systems. Based on this qualitative study, they proposed an improved model for sales chatbots. In~\cite{Gnewuch2017} they studied service quality dimensions (i.e., reliability, empathy, responsiveness, and tangibility) and how to apply them during agent design. The main difference between those works and ours is that they focus on qualitative high-level analysis while we focus on automatic detection based on the conversations logs.

\section{Methodology}
\label{methodology}

The objective of this work is to reliably detect egregious conversations between a human and a virtual agent. We treat this as a binary classification task, where the target classes are ``egregious'' and ``non-egregious''. While we are currently applying this to complete conversations (i.e., the classification is done on the whole conversation), some of the features examined here could likely be used to detect egregious conversations as they were unfolding in real time.
To perform egregious conversation detection, features from both customer inputs and agent responses are extracted, together with features related to the combination of specific inputs and responses.
In addition, some of these features are \textit{contextual}, meaning that they are dependent on \textit{where} in the conversation they appear.

Using this set of features for detecting egregious conversations is novel, and as our experimental results show, improves performance compared to a model based solely on features extracted from the conversation's text. We now describe the agent, customer, and combined customer-agent features.

\subsection{Agent Response Features}
A virtual agent is generally expected to closely simulate interactions with a human operator~\cite{Reeves1996,Nass2000,Kramer:2008}.
 When the agent starts losing the context of a conversation, fails in understanding the customer intention, or keeps repeating the same responses, the illusion of conversing with a human is lost and the conversation may become extremely annoying. With this in mind, we now describe the analysis of the agent's responses and associated features (summarized in the top part of Table~\ref{tab:setFeatures}).

\subsubsection{Repeating Response Analysis}
\label{sec:rep_res}
As typically implemented, the virtual agent's task is to reliably detect the intent of each customer's utterance and respond meaningfully. Accurate intent detection is thus a fundamental characteristic of well-trained virtual agents, and incorrect intent analysis is reported as the leading cause of user dissatisfaction~\cite{Sarikaya2017}. Moreover, since a classifier (e.g., SVM, neural network, etc.) is often used to detect intents, its probabilistic behavior can cause the agent to repeat the same (or semantically similar) response over and over again, despite the user's attempt to rephrase the same intent.

Such agent repetitions lead to an unnatural interaction~\cite{Tina2011}. To identify the agent's repeating responses, we measured similarity between agent's subsequent (not necessarily sequential) turns. We represented each sentence by averaging the pre-trained embeddings\footnote{https://code.google.com/archive/p/word2vec} of each word in the sentence, calculating the cosine similarity between the representations.
Turns with a high similarity value\footnote{Empirically, similarity values $\geq{0.8}$\label{xx}} are considered as repeating responses.

\subsubsection{Unsupported Intent Analysis}
\label{sec:unsporped}
Given that the knowledge of a virtual agent is necessarily limited, we can expect that training would not cover all customer intents. If the classifier technology provides an estimate of classification confidence, the agent can respond with some variant of ``I'm not trained on that'' when confidence is low. In some cases, customers will accept that not all requests are supported. In other cases, unsupported intents can lead to customer dissatisfaction~\cite{Sarikaya2017}, and cascade to an egregious conversation (as discussed below in Section~\ref{sec:cai}). We extracted the possible variants of the unsupported intent messages directly from the system, and later matched them with the agent responses from the logs.

\subsection{Customer Inputs Features}
From the customer's point of view, an ineffective interaction with a virtual agent is clearly undesirable. An ineffective interaction requires the expenditure of relatively large effort from the customer with little return on the investment~\cite{Zeithaml,BENMIMOUN2012605}.
These efforts can appear as behavioral cues in the customer's inputs, and include emotions, repetitions, and more. We used the following customer analysis in our model. Customer features are summarized in the middle part of Table~\ref{tab:setFeatures}.

\subsubsection{Rephrasing Analysis}
\label{sec:cusRep}
When a customer repeats or rephrases an utterance, it usually indicates a problem with the agent's understanding of the customer's intent. This can be caused by different reasons as described in~\cite{sano-kaji-sassano:2017:SIGDIAL}. To measure the similarity between subsequent customer turns to detect repetition or rephrasing, we used the same approach as described in Section~\ref{sec:rep_res}.
Turns with a high similarity value\textsuperscript{\ref{xx}} are considered as rephrases.

\subsubsection{Emotional Analysis}
The customer's emotional state during the conversation is known to correlate with the conversation's quality~\cite{oliver2014satisfaction}. In order to analyze the emotions that customers exhibit in each turn, we utilized the IBM Tone Analyzer service, available publicly online\footnote{https://ibm.co/2hnYkCv}. This service was trained using customer care interactions, and infers emotions such as \textit{frustration}, \textit{sadness}, \textit{happiness}. We focused on negative emotions (denoted as NEG EMO) to identify turns with a negative emotional peak (i.e., single utterances that carried high negative emotional state), as well as to estimate the aggregated negative emotion throughout the conversation (i.e., the averaged negative emotion intensity).
In order to get a more robust representation of the customer's negative emotional state, we summed the score of the negative emotions (such as \textit{frustration}, \textit{sadness}, \textit{anger}, etc.) into a single negative sentiment score (denoted as NEG SENT).
Note that we used the positive emotions as a filter for other customer features, such as the rephrasing analysis. Usually, high positive emotions capture different styles of ``thanking the agent'', or indicate that the customer is somewhat satisfied~\cite{RYCHALSKI201784}, thus, the conversation is less likely to become egregious.

\subsubsection{Asking for a Human Agent}
In examining the conversation logs, we noticed that it is not unusual to find a customer asking to be transferred to a human agent. Such a request might indicate that the virtual agent is not providing a satisfactory service. Moreover, even if there are human agents, they might not be available at all times, and thus, a rejection of such a request is sometimes reasonable, but might still lead to customer frustration~\cite{amsel_1992}.

\subsubsection{Unigram Input}
In addition to the above analyses, we also detected customer turns that contain exactly one word. The assumption is that single word (unigram) sentences are probably short customer responses (e.g., no, yes, thanks, okay), which in most cases do not contribute to the egregiousness of the conversation. Hence, calculating the percentage of those turns out of the whole conversation gives us another measurable feature.

\subsection{Customer-Agent Interaction Features}
\label{sec:cai}
We also looked at features across conversation utterance-response pairs in order to capture a more complete picture of the interaction between the customer and the virtual agent. Here, we considered a pair to be customer utterance followed by an agent response. For example, a pair may contain a turn in which the customer expressed negative emotions and received a response of ``not trained'' by the agent. In this case, we would leverage the two analyses: emotional and unsupported intent. Figure~\ref{fig:example} gives an example of this in the customer's penultimate turn.
Such interactions may divert the conversation towards becoming egregious.
These features are summarized in the last part of Table~\ref{tab:setFeatures}.

\subsubsection{Similarity Analysis}
We also calculated the similarity between the customer's turn and the virtual agent's response in cases of customer rephrasing. This analysis aims to capture the reason for the customer rephrasing. When a similarity score between the customer's turn and the agent's response is low, this may indicate a misclassified intent, as the agent's responses are likely to share some textual similarity to the customer's utterance.  Thus, a low score may indicate a poor interaction, which might lead the conversation to become egregious.
Another similarity feature is between two customer's subsequent turns when the agent's response was ``not trained''.

\begin{table}[t]
\resizebox{1.0\columnwidth}{!}{
    \begin{tabular}{c| l |p{6cm}|  c}
  \hline\hline \textbf{Group}&\textbf{Feature} & \textbf{Description} & \textbf{Contextual?} \\\hline
  \multirow{4}{*}{Agent}& AGNT RPT & Similarity of subsequent agent responses & Yes \\
&\#AGNT !TRND& Number of times the agent replied with ``not trained'' & No \\\hline \hline
 \multirow{4}{*}{Customer}
&   MAX 3 RPHRS & Max rephrasing similarity score of 3 subsequent turns & Yes \\
 &\#RPHRS & Number of customer rephrasing throughout the conversation & Yes \\
   &MAX NEG EMO & Max negative emotion in the conversation & No \\
&NEG SENT & Aggregated negative sentiment in the conversation & No \\
 &DIFF NEG SENT & Difference between max turn-level negative sentiment and conversation-level & Yes + No \\
 & RPHRS \& NEG SENT & Rephrasing of subsequent turns with an average high negative sentiment &Yes \\
  &HMN AGT \& NEG SENT& Negative sentiment when asking for a human agent & No \\
  &\#$1$ WRD & Turns that contained only one word & Yes \\\hline  \hline
  \multirow{4}{*}{Customer-}&NEG SENT \& AGNT !TRND& Customer negative sentiment with agent replying ``not trained'' & No \\
\multirow{4}{*}{Agent}  &HMN AGT  \& AGNT !TRND & Customer asking to talk to a human agent followed by the agent replying ``not trained'' & No \\
 \multirow{4}{*}{Interaction}   &LNG SNTNS \& AGNT !TRND & Customer long turn followed by an agent ``not trained'' response & No \\
  &RPHRS \& SMLR& The similarity between the customer's turn and the agent's response in case of customer rephrasing & No \\
  &RPHRS \& AGNT !TRND & The similarity between the customer's turns when the agent's response is ``not trained'' & No \\
  &CONV LEN & Total number of customer turns and agent responses & No \\\hline
 \end{tabular}}
 \caption{Features sets description.}~\label{tab:setFeatures}
\end{table}

\subsection{Conversation Egregiousness Prediction Classifier}
We trained a binary SVM classifier with a linear kernel. A feature vector for a sample in the training data is generated using the scores calculated for the described features, where each feature value is a number between [0,1]. After the model was trained, test conversations are classified by the model, after being transformed to a feature vector in the same way a training sample is transformed. The SVM classification model (denoted \textit{EGR}) outputs a label ``egregious'' or ``non-egregious'' as a prediction for the conversation.

\section{Experiments}
\label{experiment}

\subsection{Dataset}
\begin{figure}[t]
\centering
  \includegraphics[width=0.95\columnwidth]{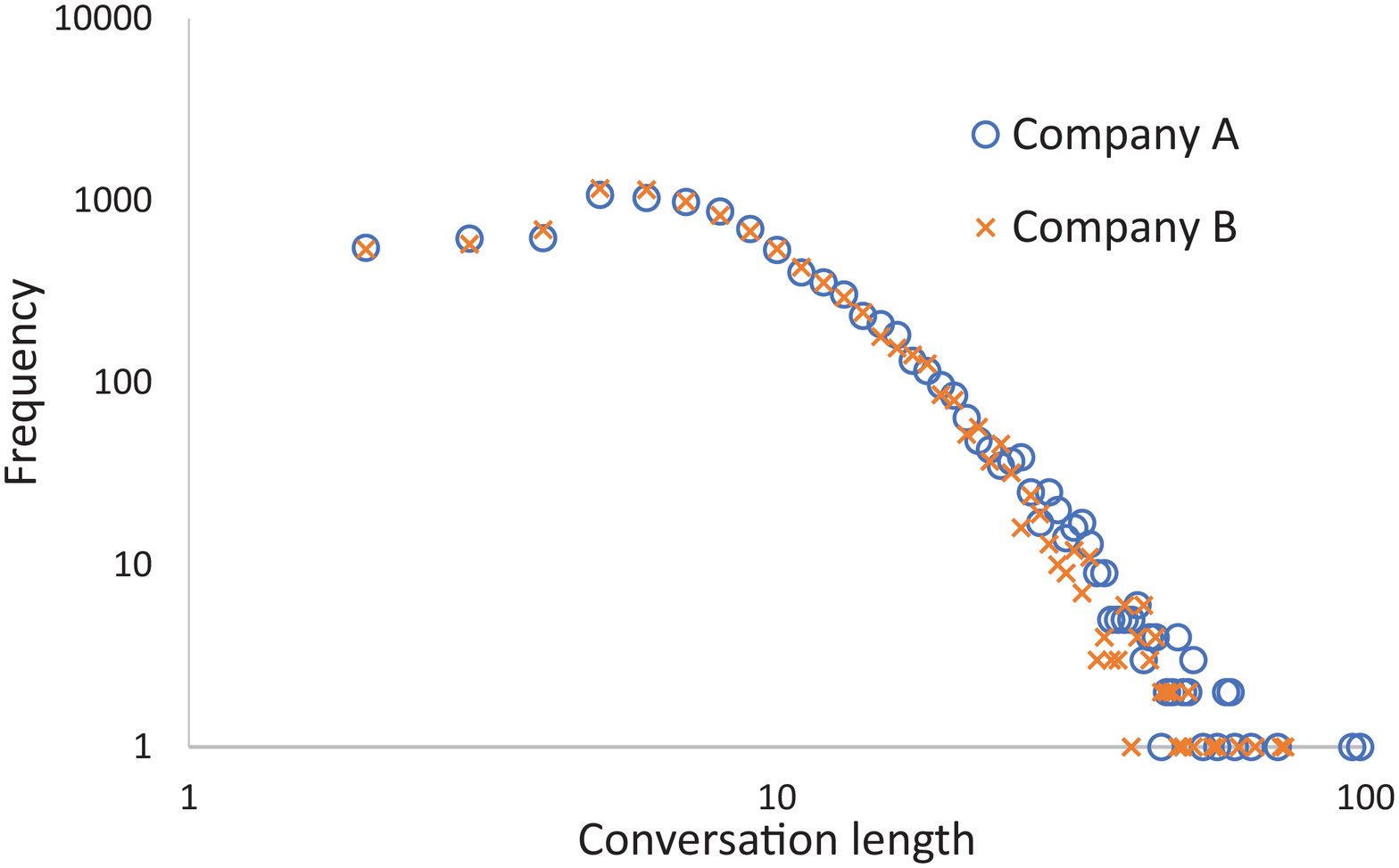}
  \caption{Frequency versus conversation length for {\sl company A} and {\sl company B} on a log-log scale.}~\label{fig:dialogLen}
\end{figure}

We extracted data from two commercial systems that provide customer support via conversational bots (hereafter denoted as {\sl company A} and {\sl company B}). Both agents are using similar underlying conversation engines, each embedded in a larger system with its own unique business logic.
{\sl Company A}'s system deals with sales support during an online purchase, while {\sl company B}'s system deals with technical support for purchased software products. Each system logs conversations, and each conversation is a sequence of tuples, where each tuple consists of \textit{\{conversation id, turn id, customer input, agent response\}}. From each system, we randomly extracted $10000$ conversations. We further removed conversations that contained fewer than $2$ turns, as these are too short to be meaningful since the customer never replied or provided more details about the issue at hand. Figure~\ref{fig:dialogLen} depicts the frequencies of conversation lengths which follow a power-law relationship. The conversations from {\sl company A}'s system tend to be longer, with an average of $8.4$ turns vs. an average of $4.4$ turns for {\sl company B}.

\subsection{Experimental Setup}
\label{GTdesc}
The first step in building a classification model is to obtain ground truth data. For this purpose, we randomly sampled conversations from our datasets. This sample included $1100$ and $200$ conversations for {\sl company A} and {\sl company B} respectively.
The sampled conversations were tagged using an in-house tagging system designed to increase the consistency of human judgements. Each conversation was tagged by four different expert judges\footnote{judges that are HCI experts and have experience in designing conversational agents systems.}. Given the full conversation, each judge tagged whether the conversation was egregious or not following this guideline: ``Conversations which are extraordinarily bad in some way, those conversations where you'd like to see a human jump in and \textit{save} the conversation''.

We generated true binary labels by considering a conversation to be egregious if at least three of the four judges agreed. The inter-rater reliability between all judges, measured by Cohen's Kappa, was $0.72$ which indicates high level agreement. This process generated the egregious class sizes of $95$ ($8.6$\%) and $16$ ($8\%$) for {\sl company A} and {\sl company B}, respectively. This verifies the unbalanced data expectation as previously discussed.

We also implemented two baseline models, rule-based and text-based, as follows:
\paragraph{Rule-based.}
In this approach, we look for cases in which the virtual agent responded with a ``not trained'' reply, or occurrences of the customer requesting to talk to a human agent. As discussed earlier, these may be indicative of the customer's dissatisfaction with the nature of the virtual agent's responses.
\paragraph{Text-based.}
A model that was trained to predict egregiousness given the conversation's text (all customer and agent's text during the conversation). This model was implemented using state-of-the-art textual features as in~\cite{HerzigSK17}. In~\cite{HerzigSK17} emotions are detected from text, which can be thought of as similar to our task of predicting egregious conversations.

We evaluated these baseline methods against our classifier using $10$-fold cross-validation over {\sl company A}'s dataset (we did not use {\sl company B}'s data for training due to the low number of tagged conversations). Since class distribution is unbalanced, we evaluated classification performance by using precision (P), recall (R) and F1-score (F) for each class. The \textit{EGR} classifier was implemented using an SVM with a linear kernel\footnote{http://scikit-learn.org/stable/modules/svm.html}.

\subsection{Classification Results}

Table \ref{results-extended} depicts the classification results for both classes and
the three models we explored. The \textit{EGR} model significantly outperformed both baselines\footnote{\textit{EGR} with $p <0.001$, using \textit{McNemar's test.}}.
Specifically, for the egregious class, the precision obtained by the text-based and \textit{EGR} models were similar. This indicates that the text analyzed by both models encodes some information about egregiousness. On the other hand, for the recall and hence the F1-score, the \textit{EGR} model relatively improved the text-based model by $41\%$ and $18\%$, respectively. We will further analyze the models below.

\begin{table}[t]
\centering
\resizebox{0.9\columnwidth}{!}{
\begin{tabular}{lcccccc}
\hline\hline
         & \multicolumn{3}{c}{Egregious}                    & \multicolumn{3}{c}{Non-Egregious}                \\
Model    & P              & R              & F              & P              & R              & F              \\\hline
Rule-based   &         0.28        &      0.54       &       0.37     &        0.95         &     0.87      &    0.91          \\
Text-based     &            0.46     &    0.56      &           0.50   &  0.96      &   \textbf{0.94}     & \textbf{0.95}  \\
EGR   &          \textbf{0.47}     &    \textbf{0.79}       &       \textbf{0.59}        &          \textbf{0.98}     &   0.92      &  \textbf{0.95}        \\\hline\hline

\end{tabular}}
\caption{Cross-validation results for the baselines and  \textit{EGR} models.}
\label{results-extended}
\end{table}

\subsection{Feature Set Contribution Analysis}
To better understand the contributions of different sets of features to our \textit{EGR} model, we examined various features in an incremental fashion. Based on the groups of feature sets that we defined in Section~\ref{methodology}, we tested the performance of different group combinations, added in the following order: \textit{agent}, \textit{customer} and \textit{customer-agent interactions}.

\begin{figure}[t]
  \includegraphics[width=0.95\columnwidth]{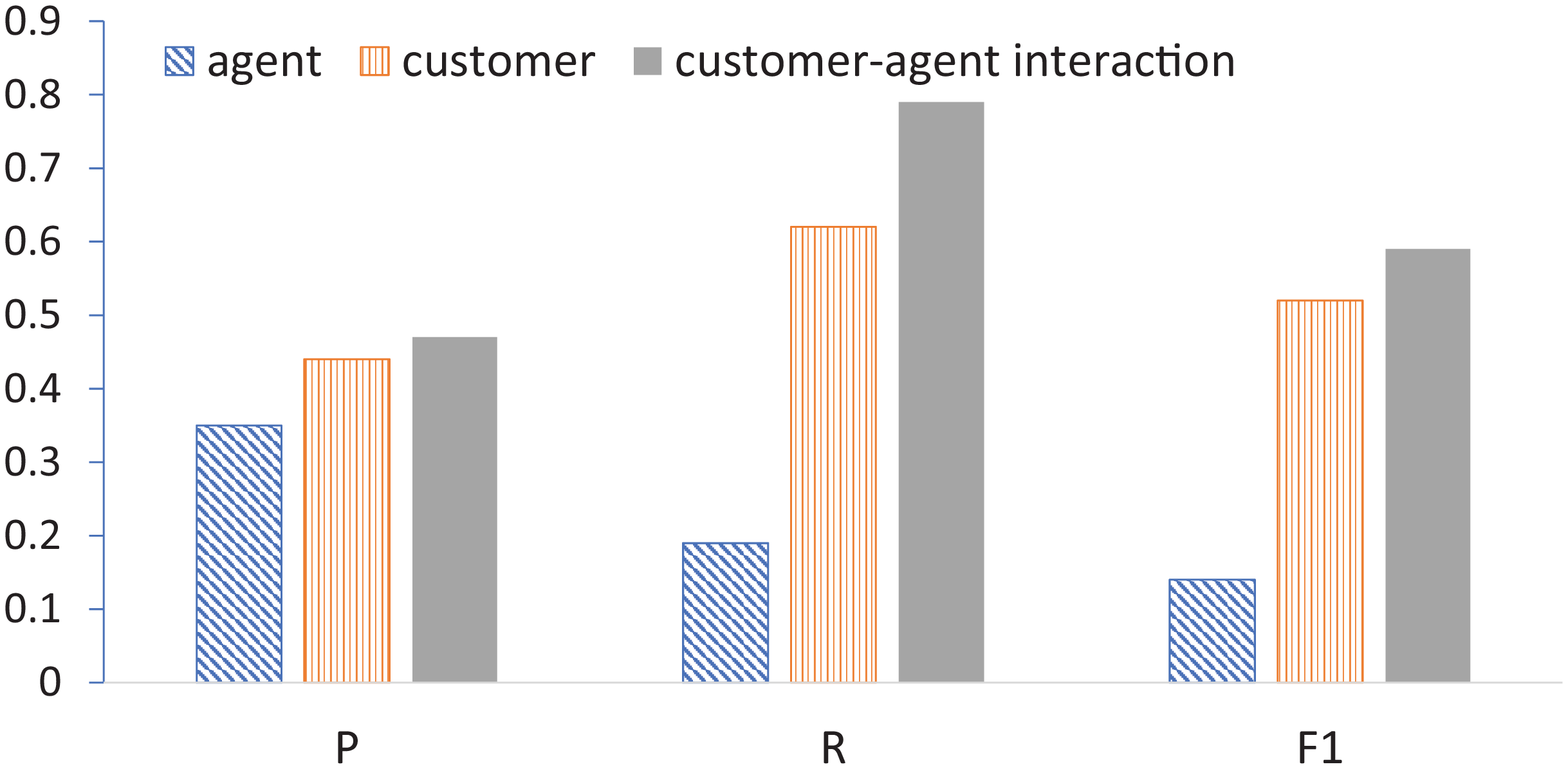}
  \caption{Precision (P), Recall (R), and F1-score (F) for various group combinations.}~\label{fig:detectionFeatures}
\end{figure}

Figure \ref{fig:detectionFeatures} depicts the results for the classification task. The $x$-axis represents specific combinations of groups, and the $y$-axis represents the performance obtained. Figure \ref{fig:detectionFeatures} shows that adding each group improved performance, which indicates the informative value of each group. The figure also suggests that the most informative group in terms of prediction ability is the \textit{customer} group.

\subsection{Cross-Domain Analysis}
We also studied how robust our features were: If our features generalize well, performance should not drop much when testing {\sl company B} with the classifier trained exclusively on the data from {\sl company A}.
Although  {\sl company A} and  {\sl company B} share similar conversation engine platforms, they are completely different in terms of objectives, domain, terminology, etc.
For this task, we utilized the $200$ annotated conversations of {\sl company B} as test data, and experimented with the different models, trained on {\sl company A}'s data. The rule-based baseline does not require training, of course, and could be applied directly.

Table~\ref{tab:cross} summarizes the results showing that the performance of the \textit{EGR} model is relatively stable (w.r.t the model's performance when it was trained and tested on the same domain),  with a degradation of only $9\%$ in F1-score\footnote{\textit{EGR} model results are statistically significant compared to the baselines models with $p <0.001$, using \textit{McNemar's test.}}. In addition, the results also show that the text-based model performs poorly when applied to a different domain (F1-score of $0.11$). This may occur since textual features are closely tied to the training domain.

\begin{table}[t]
\centering
\resizebox{0.9\columnwidth}{!}{
\begin{tabular}{lcccccc}
\hline\hline
         & \multicolumn{3}{c}{Egregious}                    & \multicolumn{3}{c}{Non-Egregious}                \\
Model    & P              & R              & F              & P              & R              & F              \\\hline
Rule-based   &         0.15        &      0.12       &       0.14   &   0.93 & 0.94 & 0.93      \\
Text-based     &             0.33     &    0.06     &           0.11 &0.92 &  \textbf{0.99} &  \textbf{0.96} \\
EGR   &         \textbf{0.41}     &    \textbf{0.81}       &       \textbf{0.54} &    \textbf{0.98} &0.90 &0.94       \\\hline\hline

\end{tabular}}
\caption{Cross domain performance ( models trained on {\sl company A}'s data, tested on {\sl company B}'s data). }
\label{tab:cross}
\end{table}

\subsection{Models Analysis}

\subsubsection{Customer Rephrasing Analysis}
Inspired by~\cite{Sarikaya2017,sano-kaji-sassano:2017:SIGDIAL} we analyzed the customer rephrasing motivations for both the egregious and the non-egregious classes. First, we detected customer rephrasing as described in Section~\ref{sec:cusRep}, and then assigned to each its motivation. Specifically, in our setting, the relevant motivations are\footnote{We did not consider other motivations like automatic speech recognition (ASR) errors, fallback to search, and backend failure as they are not relevant to our setting.}:
(1) \textit{Natural language understanding (NLU) error} - the agent's intent detection is wrong, and thus the agent's response is semantically far from the customer's turn; (2) \textit{Language generation (LG) limitation} - the intent is detected correctly, but the customer is not satisfied by the response (for example, the response was too generic); (3) \textit{Unsupported intent error} - the customer's intent is not supported by the agent.

In order to detect \textit{NLU errors}, we measured the similarity between the first customer turn (before the rephrasing) and the agent response. We followed the methodology presented in~\cite{JOVITA2015305} claiming that the best answer given by the system has the highest similarity value between the customer turn and the agent answer.
Thus, if the similarity was $<0.8$ we considered this as an erroneous detection. If the similarity was $\geq0.8$ we considered the detection as correct, and thus the rephrasing occurred due to \textit{LG limitation}.
To detect \textit{unsupported intent error} we used the approach described in Section~\ref{sec:unsporped}. As reported in table~\ref{tab:rprs}, rephrasing due to an unsupported intent is more common in egregious conversations ($18\%$ vs. $14\%$), whereas, rephrasing due to generation limitations (\textit{LG limitation}) is more common in non-egregious conversations ($37\%$ vs. $33\%$). This indicates that customers are more tolerant of cases where the system understood their intent, but the response is not exactly what they expected, rather than cases where the system's response was ``not trained''. Finally, the percentage of rephrasing due to wrong intent detection (\textit{NLU errors}) is similar for both classes, which is somewhat expected as similar underlying systems provided NLU support.

\begin{table}[t]
\centering
\resizebox{0.95\columnwidth}{!}{
\begin{tabular}{l||c|c}
\hline\hline
    & Egregious              & Non-egregious                          \\\hline
NLU error   &         $48\%$        &    $48\%$              \\
LG limitation  &            $33\%$     &   $37\%$        \\
Unsupported intent error   &        $18\%$      &  $14\%$             \\\hline\hline

\end{tabular}}
\caption{Percentage of different customer rephrasing reasons for egregious, and non-egregious conversations.}
\label{tab:rprs}
\end{table}

\subsubsection{Recall Analysis}
We further investigated why the \textit{EGR} model was better at identifying egregious conversations (i.e., its recall was higher compared to the baseline models). We manually examined $26$ egregious conversations that were identified justly so by the \textit{EGR} model, but misclassified by the other models. Those conversations were particularly prevalent with the agent's difficulty to identify correctly the user's intent due to \textit{NLU errors} or \textit{LG limitation}. We did not encounter any \textit{unsupported intent errors} leading to customer rephrasing, which affected the ability of the rule-based model to classify those conversations as egregious. In addition, the customer intents that appeared in those conversations were very diverse. While customer rephrasing was captured by the \textit{EGR} model, for the text-based model some of the intents were new (did not appear in the training data) and thus were difficult for the model to capture.

\section{Conclusions and Future Work}
\label{conclusions}
In this paper, we have shown how it is possible to detect egregious conversations using a combination of customer utterances, agent responses, and customer-agent interactional  features. As explained, the goal of this work is to give developers of automated agents tools to detect and then solve problems created by exceptionally bad conversations. In this context, future work includes collecting more data and using neural approaches (e.g., RNN, CNN) for analysis, validating our models on a range of domains beyond the two explored here. We also plan to extend the work to detect egregious conversations in real time (e.g., for escalating to a human operators), and create log analysis tools to analyze the root causes of egregious conversations and suggest possible remedies.

\balance
\bibliography{refrences}
\bibliographystyle{acl_natbib}

\end{document}